\documentclass[letterpaper, 10 pt, journal, twoside]{ieeetran}
\usepackage{amsmath,amsfonts}
\usepackage{algorithmic}
\usepackage{algorithm}
\usepackage{array} 
\usepackage{booktabs}
\usepackage{multirow}
\usepackage[caption=false,font=normalsize,labelfont=sf,textfont=sf]{subfig}
\usepackage{textcomp}
\usepackage{stfloats}
\usepackage{url}
\usepackage{verbatim}
\usepackage{graphicx}
\usepackage{cite}
\usepackage{todonotes}

\begin{document}

\title{Semantic Area Graph Reasoning for Multi-Robot 
\\ Language-Guided Search}

\author{Ruiyang Wang, Hao-Lun Hsu, Jiwoo Kim, and Miroslav Pajic}

\markboth{Journal of \LaTeX\ Class Files,~Vol.~14, No.~8, August~2021}%
{Ruiyang Wang \MakeLowercase{\textit{et al.}}: Semantic Area Graph Reasoning for Multi-Robot Language-Guided Search}


\maketitle

\begin{abstract}
Coordinating multi-robot systems (MRS) to search in unknown environments is particularly challenging for tasks that require semantic reasoning beyond geometric exploration. Classical coordination strategies rely on frontier coverage or information gain and cannot incorporate high-level task intent, such as searching for objects associated with specific room types. We propose \textit{Semantic Area Graph Reasoning} (SAGR), a hierarchical framework that enables Large Language Models (LLMs) to coordinate multi-robot exploration and semantic search through a structured semantic-topological abstraction of the environment. SAGR incrementally constructs a semantic area graph from a semantic occupancy map, encoding room instances, connectivity, frontier availability, and robot states into a compact task-relevant representation for LLM reasoning. The LLM performs high-level semantic room assignment based on spatial structure and task context, while deterministic frontier planning and local navigation handle geometric execution within assigned rooms. Experiments on the Habitat-Matterport3D dataset across 100 scenarios show that SAGR remains competitive with state-of-the-art exploration methods while consistently improving semantic target search efficiency, with up to 18.8\% in large environments. These results highlight the value of structured semantic abstractions as an effective interface between LLM-based reasoning and multi-robot coordination in complex indoor environments.
\end{abstract}


\section{Introduction}
Coordinating multi-robot systems (MRS) to accomplish complex tasks in partially observable environments is a fundamental challenge in robotics. 
%By enabling multiple robots to operate collaboratively, MRS can significantly improve efficiency, robustness, and spatial coverage compared to single-robot systems. However, 
In particualr, effectively coordinating robot teams and distributing tasks among agents remains an active research problem. %A widely adopted framework for addressing this challenge is multi-robot task allocation (MRTA), which formulates coordination as the problem of assigning robots to tasks in a way that optimizes system-level performance~\cite{gerkey2004formal, Korsah2013ACT}. 
Multi-robot task allocation (MRTA), which formulates coordination as assigning robots to tasks in a way that optimizes system-level performance, 
is a widely adopted framework~\cite{gerkey2004formal, Korsah2013ACT}.
Existing MRTA approaches span a range of coordination mechanisms, including centralized optimization-based methods that compute globally optimal assignments~\cite{burgard2005coordinated, Chakraa2023OptimizationTF} and decentralized strategies such as auction-based mechanisms where robots bid for tasks based on local cost estimates~\cite{dias2006market, zlot2006market}. %These approaches form the foundation of many multi-robot coordination systems and continue to be an important area of research in robotics.

Multi-robot coordination for exploration and search in unknown environments typically relies on geometric objectives derived from occupancy maps, such as frontier boundaries~\cite{yamauchi1997frontier} or information-gain metrics~\cite{selin2019efficient}. To distribute these spatial goals, existing methods utilize Voronoi-based partitioning~\cite{Wang2024sensor}, utility heuristics~\cite{burgard2005coordinated}, or combinatorial routing formulations (e.g., the Capacitated Vehicle Routing Problem (CVRP)~\cite{zhou2023racer}). While modern robotic systems increasingly have access to semantic information through perception and mapping pipelines that associate geometric maps with objects and scene structures~\cite{Garg2021SemanticsFR, chaplot2020object, Gervet2022NavigatingTO}, existing coordination strategies still primarily rely on geometric objectives. Consequently, they cannot effectively incorporate these semantic priors or high-level task descriptions when coordinating robot teams, often resulting in exploration strategies that lack context-aware prioritization.

Recent advances in foundation models have created new opportunities for incorporating high-level reasoning into robotic systems. Large Language Models (LLMs) and Vision-Language Models (VLMs) have demonstrated strong capabilities in multimodal reasoning, semantic grounding, and long-horizon decision making across perception and action domains. Systems such as SayCan~\cite{ahn2022can}, VIMA~\cite{jiang2023vima}, and RT-2~\cite{zitkovich2023rt} illustrate that foundation models can interpret natural language instructions and generate structured action plans for robotic tasks. Motivated by these capabilities, recent works have begun exploring the use of LLMs as high-level planners for embodied multi-robot coordination~\cite{yu2023co, shen2025enhancing, Wang2025COMRESVLMCM}. However, these existing frameworks rely on forwarding high-dimensional visual observations or raw map images to the model for reasoning. Such dense representations significantly increase prompt size and computational cost, limiting their applicability for real-time decision making in real-world MRS.

In this work, we introduce \textit{\textbf{Semantic Area Graph Reasoning (SAGR)}}, a hierarchical framework that enables LLM-based coordination through structured topological abstractions for multi-robot exploration and semantic search while maintaining practical real-time performance. Rather than reasoning directly over dense visual inputs, SAGR constructs a compact \emph{semantic area graph} that represents the discovered environment as a set of room-level nodes connected through spatial adjacency relationships. Each node encodes semantic room information together with frontier statistics and robot states. This abstraction captures the structural and semantic organization of the environment while drastically reducing the dimensionality of the information provided to the LLM.

The %proposed 
SAGR framework integrates three complementary layers: a \emph{semantic reasoning layer}, in which an LLM assigns robots to room instances by reasoning over a semantic area graph; a \emph{frontier selection layer}, which allocates frontier clusters within the assigned rooms; and a \emph{local execution layer}, which performs motion planning and sensing. The key contribution is not merely the use of an LLM for coordination, but the introduction of a \emph{structured semantic area graph} that serves as the interface between semantic decision-making and geometric execution. This abstraction compresses the incrementally discovered environment into task-relevant entities that preserve room-level semantics, topology, frontier availability, and robot occupancy, enabling efficient high-level reasoning without operating directly on dense maps or raw visual inputs. By combining this structured abstraction with deterministic frontier-based planning, SAGR retains the efficiency of classical exploration methods while enabling task-aware multi-robot coordination guided by semantic context.

We evaluate the %proposed 
SAGR framework on realistic indoor environments from the Habitat-Matterport3D (HM3D) dataset~\cite{ramakrishnan2021hm3d}. %Results 
We show that, while specialized geometric strategies remain strong baselines for pure exploration, SAGR preserves competitive exploration performance while substantially improving efficiency on semantic search tasks that require task-aware coordination. Importantly, the semantic area graph keeps coordination prompts compact by abstracting the discovered environment into task-relevant room-level entities, enabling near real-time LLM inference for %multi-robot 
MRS coordination.

The main contributions are summarized as follows:
\begin{itemize}
    \item We introduce a \textbf{semantic area graph} representation that compresses dense semantic occupancy maps into a compact room-level semantic graph, preserving task relevant information including room semantics, spatial connectivity, frontier availability, and robot occupancy for efficient structured reasoning.

    \item We develop %\textbf{Semantic Area Graph Reasoning (SAGR)}, 
    SAGR, a hierarchical framework that uses~the semantic area graph %as the 
    to interface between high-level LLM-based semantic decision-making and low-level frontier planning for %multi-robot 
    MRS exploration and semantic~search.

    \item We demonstrate through extensive experiments in complex indoor environments that the proposed abstraction enables efficient multi-robot coordination, remains competitive with state-of-the-art exploration baselines, and consistently improves semantic target search efficiency.
\end{itemize}

\section{Related Work}
\subsection{Exploration and Search}
Autonomous exploration in unknown environments has been extensively studied in robotics, particularly in the context of simultaneous localization and mapping (SLAM) and information-driven planning. Early work introduced frontier based exploration, where robots iteratively navigate to boundaries between explored and unexplored regions to expand map coverage~\cite{yamauchi1997frontier}. 
%Subsequent research extended this paradigm using 
This was %later 
extended with %the use of 
information formulations that select exploration targets maximizing expected information gain~\cite{bourgault2002information, stachniss2004exploration}. Sampling-based and receding-horizon planners further improved exploration efficiency by selecting informative viewpoints in large environments~\cite{charrow2015information, bircher2016receding}. 

Exploration has also been studied in %multi-robot 
MRS settings, where coordination mechanisms distribute exploration tasks across robots while minimizing redundant coverage %. Foundational work on multi-robot task allocation formalized coordination problems in multi-agent systems~
\cite{gerkey2004formal}. Market-based approaches allow robots to bid for tasks based on estimated utility~\cite{dias2006market, zlot2006market}, while assignment-based methods compute optimal allocations using algorithms such as the Hungarian method~\cite{kuhn1955hungarian}. In exploration contexts, coordinated frontier assignment strategies distribute exploration targets across robots to improve coverage efficiency~\cite{burgard2005coordinated}. 

% More recent work has explored 
More recently, hierarchical and globally coordinated exploration frameworks %to improve 
have been shown to improve scalability in large environments; e.g., %Wang \emph{et al.}~\cite{Wang2024sensor} uses
Dynamic Voronoi cells (DVC) are used for MRS task allocation~\cite{Wang2024sensor}, while RACER~\cite{zhou2023racer} introduces a hierarchical grid-based exploration framework with CVRP for collaborative exploration. Despite these advances, most existing MRS exploration methods rely primarily on geometric heuristics or information gain metrics and lack mechanisms to incorporate high-level semantic reasoning when searching for task-specific objects or regions.

\subsection{Semantic Understanding and Mapping}
Semantic understanding of environments has become increasingly important in robotics, as many %task-oriented behaviors 
tasks require robots to reason not only about free space and obstacles, but also about %objects, places, and 
scene context. Early semantic mapping frameworks augmented spatial maps with symbolic representations of objects and places~\cite{galindo2005multi, pronobis2010multi}. Subsequent surveys further highlighted the importance of integrating semantic knowledge into robot perception, mapping, and planning pipelines~\cite{kostavelis2015semantic, crespo2020semantic}.

% Recent advances 
Advances in perception and mapping have made such semantic information increasingly accessible in modern robotic systems. Semantic SLAM frameworks incorporate object detection and scene understanding into mapping pipelines to construct enriched environment representations~\cite{bowman2017probabilistic, mccormac2017semanticfusion, rosinol2020kimera}. Higher-level abstractions such as scene graphs and topological semantic representations further capture relationships among objects, places, and spatial structure~\cite{rosinol20203d, garg2024robohop}. Together, %these works show that 
robots can now obtain semantic information alongside geometry, providing a basis for task-oriented reasoning beyond purely metric exploration. Yet, to the best of our knowledge, using scene graphs for MRS exploration and search in unknown environments remains relatively underexplored, %thus motivating the proposed method.
motivating this~work. 

\subsection{Foundation Models in Robotics}
Recent advances in foundation models have introduced new opportunities for integrating high-level reasoning into robotic systems. %Large Language Models (LLMs) and Vision-Language Models (VLMs) 
LLMs and VLMs have demonstrated strong capabilities in multimodal reasoning across perception, language understanding, and decision making. Early work explored grounding language instructions in robotic planning frameworks~\cite{ahn2022can}. More recent systems such as VIMA~\cite{jiang2023vima}, RT-2~\cite{zitkovich2023rt}, and PaLM-E~\cite{Driess2023PaLMEAE} demonstrate that VLMs can integrate perception and language reasoning for robotic control.

These capabilities have motivated growing interest in applying foundation models to embodied agents and navigation tasks. Vision-language navigation frameworks study how agents interpret language instructions to navigate complex environments~\cite{Anderson2018VisionandLanguageNI, Gu2022VisionandLanguageNA}. LLM-based planning methods further demonstrate that language models can generate executable policies and task plans %for embodied agents~ 
\cite{Huang2022LanguageMA, liang2023code}. Recent work has also begun exploring foundation model based coordination for MRS~\cite{yu2023co, shen2025enhancing, Wang2025COMRESVLMCM}. However, existing foundation-model based frameworks primarily reason over high-dimensional visual observations or dense occupancy maps, leading to large prompt sizes and significant computational overhead. More importantly, they lack a structured coordination abstraction that explicitly preserves the task-relevant semantic and topological organization of the environment for multi-robot decision-making. In contrast, the %proposed 
SAGR framework introduces a semantic area graph that abstracts the discovered environment into room-level entities with associated semantics, connectivity, frontier availability, and robot occupancy. This compact semantic-topological representation serves as the interface between high-level LLM reasoning and low-level geometric execution, enabling efficient and task-aware coordination for multi-robot exploration and semantic search in complex indoor environments.

\section{Problem Definition}
We consider %a multi-robot system (MRS) 
an MRS consisting of $M$ robots deployed in a previously unknown indoor environment $\Omega \subset \mathbb{R}^3$. Each robot $i$ operates on the ground plane with state $X_i^t = [x_i^t \; y_i^t \; \theta_i^t]^T \in \mathbb{R}^3$, where $(x_i^t, y_i^t)$ denotes planar position ($P_i^t = [x_i^t \; y_i^t]^T$) and $\theta_i^t$ the heading at time $t$.  %Let $P_i^t = [x_i^t \; y_i^t]^T \in \mathbb{R}^2$ denote the planar position of robot $i$. 
Robots must satisfy collision avoidance constraints $\|P_i^t - P_j^t\|_2 \ge d_{\text{safe}}$ for all $i \neq j$ and all $t$, and velocity constraints $\|P_i^{t+1} - P_i^t\|_2 \le V_i^{\max}$.

Each robot is equipped with a single onboard camera with field-of-view (FoV) characterized by maximum sensing range $d_{\text{det}}$ and angular range $\theta_{\text{det}}$. At timestep $t$, robot $i$ observes all non-occluded regions within its FoV. The perception module provides geometric occupancy information together with estimated semantic room-type labels inferred from visual observations using standard scene recognition models~\cite{zhou2017places}. %In this work, these 
These semantic labels are treated as outputs of an upstream perception system, and our focus is on multi-robot coordination based on such semantic estimates rather than semantic classification itself.

The environment is incrementally reconstructed as a partially observable grid map $\mathcal{G}^t \in \mathbb{Z}^{H \times W}$ with fixed spatial resolution $r$, where $H$ and $W$ denote the map height and width. Each cell $g \in \mathcal{G}^t$ represents a discrete spatial region, % and is 
defined as $g = (o(g), s(g))$, where $o(g) \in \{\textit{unknown}, \textit{free}, \textit{occupied}\}$ denotes the occupancy state and $s(g) \in \mathcal{S}$ denotes the current semantic room-label estimate for the cell when/if such semantic information is available.
These occupancy states partition the environment into three disjoint subspaces: the free space $\mathcal{G}^t_{fr}$, the occupied space $\mathcal{G}^t_{oc}$, and the unknown space $\mathcal{G}^t_{un}$ at timestep $t$, such that
$
\mathcal{G}^t = \mathcal{G}^t_{fr} \cup \mathcal{G}^t_{oc} \cup \mathcal{G}^t_{un}$ for all $t$.

Although \emph{semantic target search} is the primary task considered in this work, pure exploration is also evaluated as a baseline setting. In practice, our method naturally reduces to an exploration strategy when no instance of the target room type has yet been discovered or when no prior information about the target room is provided. Thus, we consider two subtasks: %task settings:

\paragraph{Exploration}
% Let $\mathcal{G}^t_{fr}$ and $\mathcal{G}^t_{oc}$ denote the sets of discovered free and occupied cells at timestep $t$, and let 
Let $\mathcal{G}_{res}$ denote permanently inaccessible regions. Exploration is considered complete when
\[
|\mathcal{G}^t_{fr} \cup \mathcal{G}^t_{oc}| \ge \pi_{\text{threshold}} |\mathcal{G} \setminus \mathcal{G}_{res}|,
\]
where $\pi_{\text{threshold}} \in [0,1]$ specifies the required coverage ratio.

\paragraph{Semantic Target Search}
The objective is to locate a target object at an unknown cell $g_{\text{target}} \in \mathcal{G}$. We assume prior knowledge that the object is associated with a semantic room type $s_{\text{target}} \in \mathcal{S}$. Because the environment may contain multiple instances of this room type and the specific instance containing the target is unknown, robots must first explore the environment to discover room instances and their semantic labels, then prioritize search among cells satisfying $s(g)=s_{\text{target}}$. The task is complete once the target cell is observed.

\section{The SAGR Approach}

\begin{figure}
    \centering
    \includegraphics[width=1.0\linewidth]{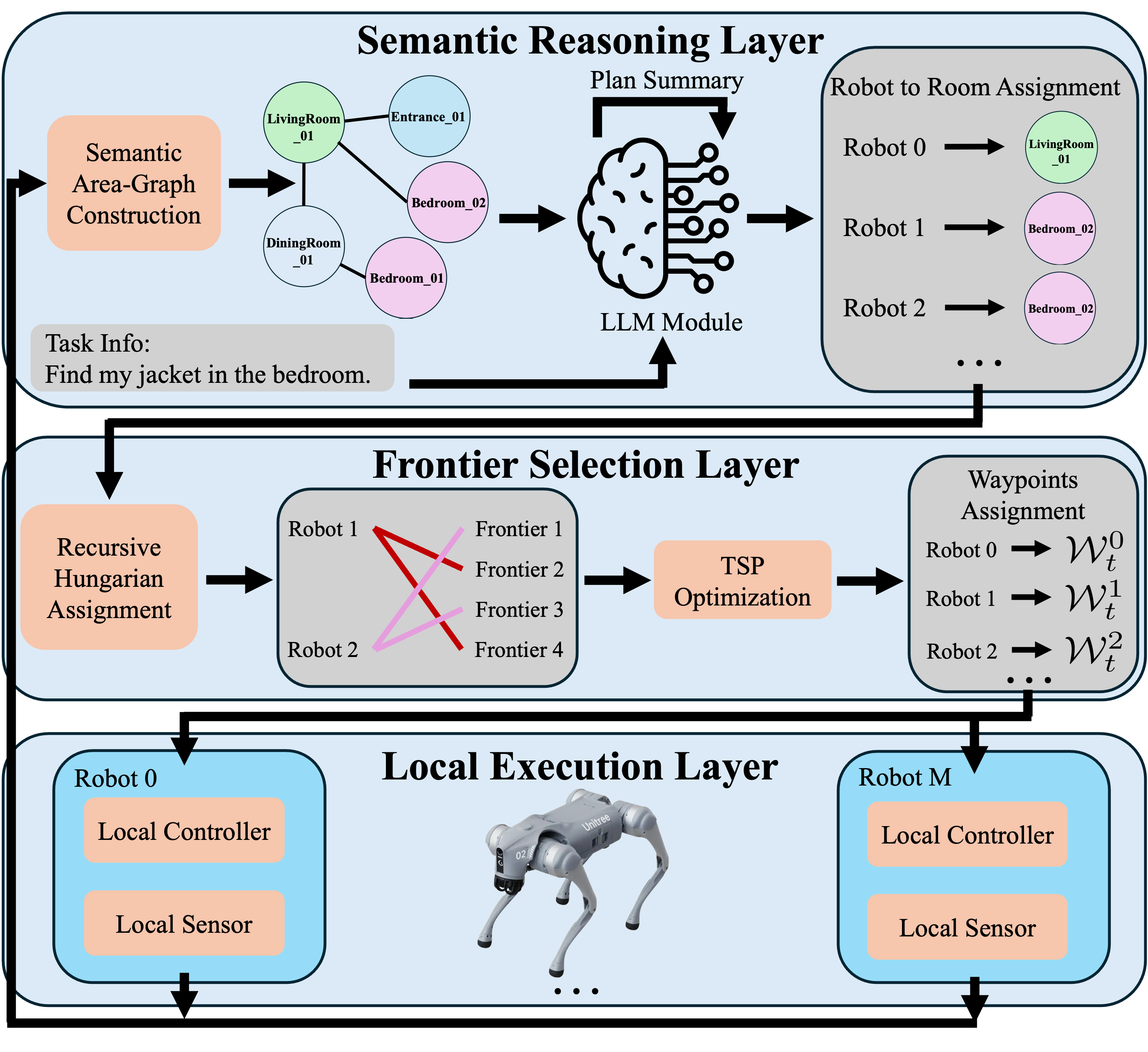}
    \caption{Overview of the SAGR framework. The semantic reasoning layer assigns robots to rooms using an LLM operating on the semantic area graph. The frontier selection layer allocates frontier clusters and optimizes waypoint order, while the local execution layer performs navigation and sensing.}
    \label{fig:pipeline}
\end{figure}

% We propose \textbf{SAGR} (Semantic Area-Graph Reasoning for Multi-Robot Language-Guided Search), 
\textbf{SAGR} is a hierarchical framework for multi-robot exploration and semantic target search that integrates semantic room-level reasoning with classical frontier-based exploration. The key idea is to elevate coordination from cell-level frontier assignment to room-level semantic reasoning while retaining deterministic optimization for low-level motion planning. At each coordination cycle, SAGR reconstructs a semantic area graph from the current semantic occupancy map provided by the upstream perception module and uses this graph as the structured coordination state for high-level decision-making. This semantic area graph serves as the interface between LLM-based semantic reasoning and low-level geometric execution.

As illustrated in Fig.~\ref{fig:pipeline}, SAGR consists of three layers: i) a \textit{semantic reasoning layer} that assigns robots to candidate room instances based on the semantic area graph together with prior task information, when available; ii) a \textit{frontier selection layer} that determines frontier waypoints within the assigned rooms; iii) a \textit{local execution layer} that performs motion control and sensing. High-level coordination is executed every $T_{\text{coord}}$ steps. Between coordination events, robots continue executing the frontier waypoints assigned during the most recent planning.

\subsection{Frontier Detection}

Given the semantic occupancy grid $\mathcal{G}^t$, we extract frontier cells following the classical frontier exploration %formulation~
\cite{yamauchi1997frontier}. A free cell $g \in \mathcal{G}^t_{fr}$ is defined as a frontier if it is adjacent to at least one unknown cell:
$$
g \in \mathcal{G}^t_{fr}
\quad \text{and} \quad
\exists g' \in \mathcal{N}_4(g)
\text{ such that }
g' \in \mathcal{G}^t_{un},
$$
where $\mathcal{N}_4(g)$ denotes the 4-connected neighborhood of $g$.

Frontier cells represent boundaries between explored and unexplored regions and therefore define candidate expansion locations. To obtain spatially meaningful targets, frontier cells are grouped into clusters using breadth-first search (BFS) over free-space connectivity. Clustering is restricted by a maximum BFS distance threshold $c_{\max}$ to prevent excessive merging.

Each frontier cluster $f_i \in \mathcal{F}$ is represented by a representative waypoint $p^i_{rep}$ and a cluster size $c^i_{size}$; i.e.,
$
f_i = \{p^i_{rep}, c^i_{size}\}.
$
The representative waypoint $p^i_{rep}$ is computed as the arithmetic centroid of the cluster and then projected to the nearest reachable free cell in the observed map. Each frontier cluster is then associated with a room instance based on the semantic label of the room instance containing $p^i_{rep}$. These frontier-to-room associations are stored as node attributes in the semantic area graph described next.

\subsection{Semantic Area Graph Construction}

To capture the discovered semantic structure of the environment, we construct a semantic area graph
$
\mathcal{A}^t = (\mathcal{V}^t, \mathcal{E}^t),
$
at each coordination cycle, where nodes correspond to discovered room instances and edges encode spatial adjacency between them in the currently observed map.

Room instances are extracted from the semantic occupancy grid $\mathcal{G}^t$. For each room type $s \in \mathcal{S}$, we identify connected components of cells satisfying $o(g)=\textit{free}$ and $s(g)=s$ using 4-connected grid connectivity. Here, a \emph{room type} refers to a semantic label in $\mathcal{S}$ (e.g., bedroom or kitchen), whereas a \emph{room instance} refers to one connected component of free cells with that semantic label. Each connected component forms a distinct room instance and is represented as a node $v \in \mathcal{V}^t$ with a unique identifier $\text{room\_id}(v)$. Consequently, multiple nodes may share the same room type while corresponding to spatially disconnected room instances.

\begin{figure}
    \centering
    \includegraphics[width=1.0\linewidth]{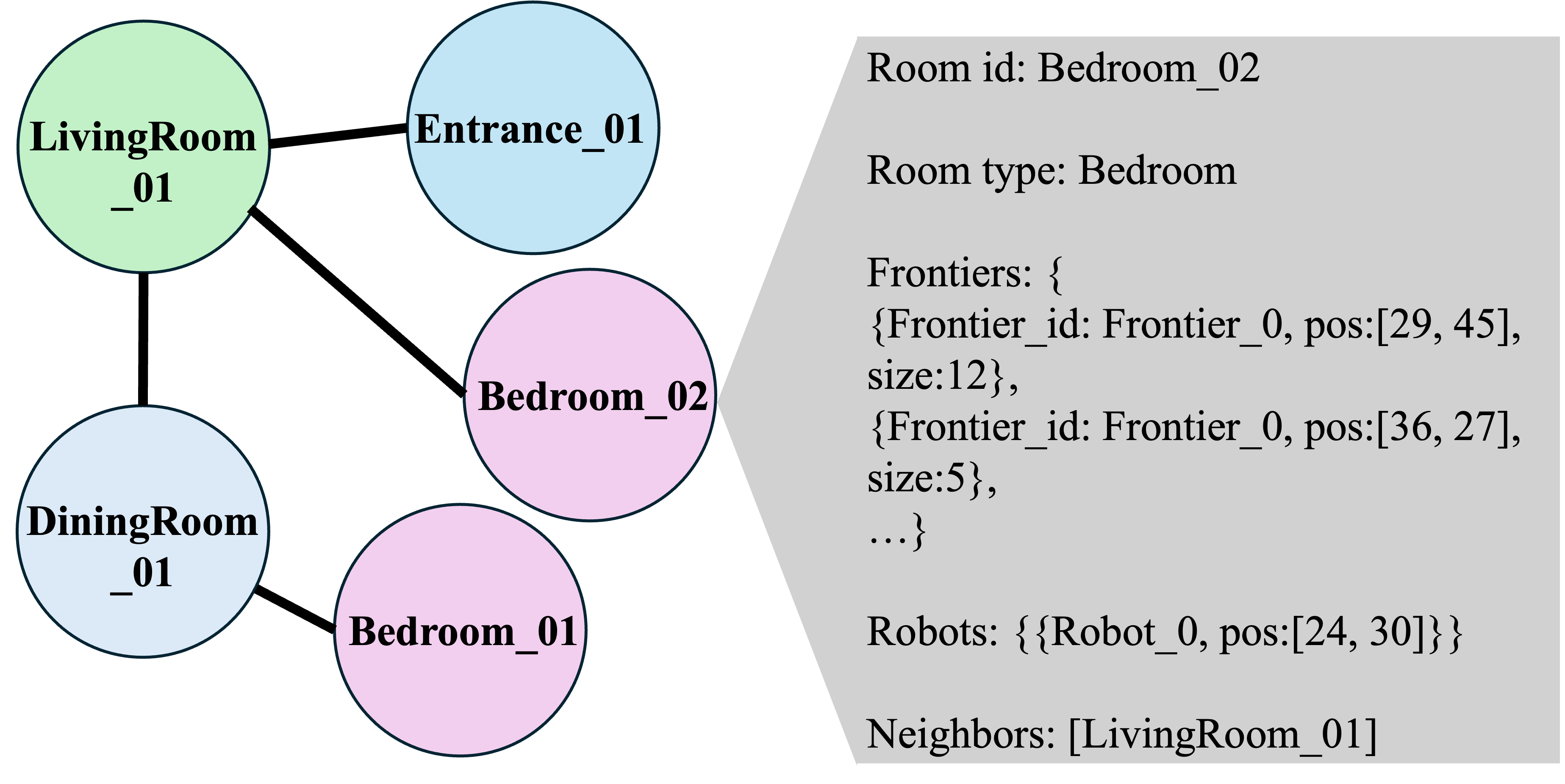}
    \caption{Example semantic area graph constructed from the observed semantic map. Nodes represent discovered room instances and edges represent spatial adjacency. Each node stores room attributes including frontier clusters, robots, and neighboring rooms.}
    \label{fig:graph}
\end{figure}

Each node $v \in \mathcal{V}^t$ stores the attributes as shown in Fig.~\ref{fig:graph}. An edge $(v_i, v_j) \in \mathcal{E}^t$ is created when two distinct room instances contain 4-connected adjacent free cells in the currently observed map. In this way, adjacency is recomputed at each coordination cycle as new free space is observed, allowing the graph structure to evolve together with the discovered environment. Room nodes that contain no frontier clusters are considered fully explored and removed from the semantic area graph. This pruning step keeps the graph compact and ensures that the LLM focus on rooms that contain unexplored space.

This representation converts the cell-level semantic map into a semantic-topological graph, enabling coordination over room instances instead of individual frontier cells while preserving spatial relationships and reducing the reasoning state for the LLM. Fig.~\ref{fig:graph} illustrates an example semantic area graph and the information stored in each node.

\subsection{LLM-Based Room-Level Planning}
At each coordination cycle, the semantic area graph $\mathcal{A}^t$ is serialized into a structured textual representation and provided to a foundation LLM for high-level coordination. The serialized graph includes, for each room instance, its room identifier, room type, associated frontier information, currently assigned robots, and neighboring room identifiers.

As in Fig.~\ref{fig:pipeline}, the LLM receives the following information:
\begin{itemize}
    \item the serialized semantic area graph,
    \item the previous plan summary,
    \item the task specification,
    \item the target room type $s_{\text{target}}$ for semantic search tasks.
\end{itemize}

For exploration tasks, the LLM assigns robots across discovered room instances while favoring rooms with greater frontier availability %so as 
to improve global exploration coverage. For semantic target search tasks, the LLM prioritizes room instances whose semantic type matches the target room type $s_{\text{target}}$, while maintaining spatial diversity across robots to reduce redundant search.

The plan summary provides a compact description of the coordination strategy from the preceding cycle, which helps maintain temporal consistency as the graph evolves and reduces oscillation in room assignments across replanning steps.

Based on this intomation, the LLM produces two outputs:
\begin{enumerate}
    \item a short plan summary describing the current high-level coordination strategy;
    \item a structured mapping from each robot to exactly one room identifier, where multiple robots may be assigned to the same room instance when appropriate.
\end{enumerate}
This room-assignment output is then passed to the downstream frontier selection layer.

\subsection{Frontier Selection}

Given the room assignments produced by the LLM, each robot restricts its exploration to the frontier clusters associated with its assigned room instance. Let $\mathcal{F}_v$ denote the set of frontier clusters associated with room instance $v$.

If multiple robots are assigned to the same room, % instance, 
the frontier clusters in $\mathcal{F}_v$ are distributed among them to reduce redundant coverage. This allocation is formulated as a bipartite matching problem between the robots assigned to room $v$ and the frontier representative waypoints in $\mathcal{F}_v$. The matching cost is defined by the Euclidean distance between each robot's current position and each frontier representative location $p^i_{rep}$, and the assignment is solved using the Hungarian algorithm.

Because the number of frontier clusters may exceed the number of robots assigned to a room, the matching procedure is applied iteratively to construct a multi-frontier allocation. At each iteration, the Hungarian algorithm assigns at most one frontier cluster to each robot based on the current cost matrix. The assigned frontier clusters are then removed from $\mathcal{F}_v$, and the matching is repeated until all frontier clusters in the room have been allocated. For each robot, the resulting sequence of assigned frontier representatives is then reordered using a Traveling Salesman Problem (TSP) solver to reduce traversal distance within the room.

This layer enables efficient local exploration within each assigned room instance while reducing redundant coverage when multiple robots operate in the same room.

\subsection{Local Execution}
Between coordination events, each robot executes the sequence of frontier waypoints assigned during the frontier selection stage. Robots navigate to these frontier representatives sequentially using motion planning on the current observed free-space map. If a robot exhausts its assigned frontier clusters or encounters an infeasible path due to newly discovered obstacles or map changes, it falls back to the nearest currently reachable frontier in the map to maintain continuous exploration.

% Every $T_{\text{coord}}$ steps, the semantic area graph $\mathcal{A}^t$ is reconstructed from the updated semantic occupancy map and the full coordination process is repeated. The newly generated room assignments and frontier queues replace the previous ones, allowing the system to adapt online as new rooms, semantic labels, and frontier regions are discovered during exploration.

\section{Evaluations}
\begin{figure}
    \centering
    \includegraphics[width=0.92\linewidth]{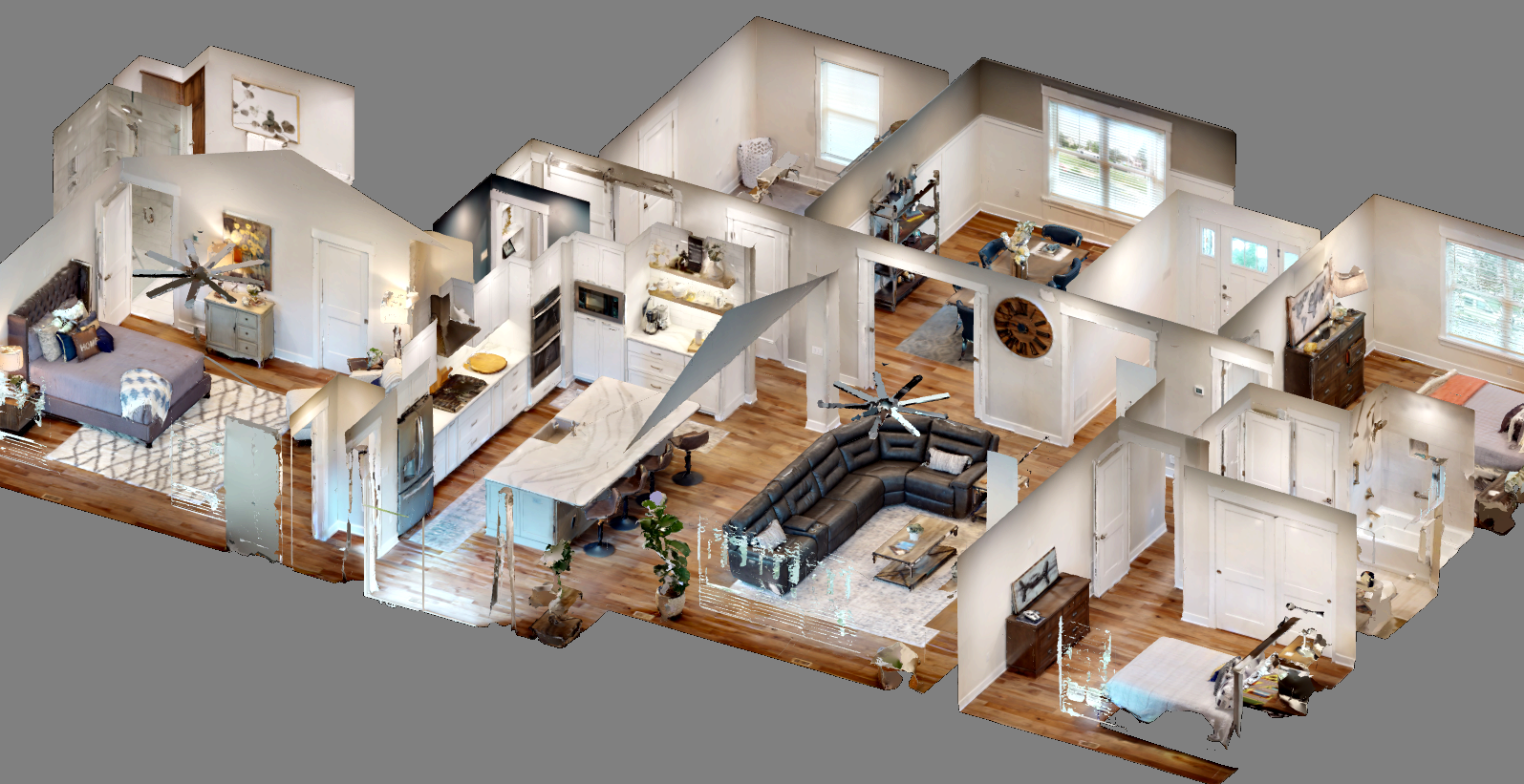}
    \caption{Example environment from the HM3D dataset, showing an indoor apartment scene with semantic room layouts used for our search experiments.}
    \label{fig:apartment}
\end{figure}

To evaluate the effectiveness of the proposed algorithm, we conduct extensive experiments on the %Habitat-Matterport 3D (HM3D) 
HM3D dataset~\cite{ramakrishnan2021hm3d}, which provides realistic indoor apartment layouts with furniture and semantic room annotations (see e.g., Fig.~\ref{fig:apartment}). %An example environment is shown in Fig.~\ref{fig:apartment}. 
We select 10 different apartment layouts. For each layout, we generate 10 scenarios by randomly sampling robot initial poses (positions and orientations) and target object locations within the environment. In total, we test 100 different scenarios.

\subsection{Baselines}
% To evaluate the proposed method, we 
We compare SAGR against several representative geometric coordination strategies for %multi-robot 
MRS exploration in unknown environments. Although these baselines are not designed to explicitly exploit semantic room-type priors, they provide strong baselines for assessing how much semantic reasoning improves coordination performance.

\textbf{i) Hungarian Frontier Assignment.}  
This classical approach assigns frontier clusters to robots based on Euclidean distance using the Hungarian algorithm~\cite{kuhn1955hungarian}. After assignment, each robot computes an efficient traversal order over its assigned frontier clusters using a TSP solver. 

\textbf{ii) RACER.}  
RACER~\cite{zhou2023racer} is a recent state-of-the-art multi-robot exploration method that discretizes the environment into hierarchical grids (HGrid) based on unknown-cell ratios. Multi-robot coordination is then formulated as a pairwise CVRP to distribute exploration tasks.

\textbf{iii) AEP + DVC.}  
AEP~\cite{selin2019efficient} is a state-of-the-art method for single-robot exploration that samples candidate viewpoints based on expected information gain. To extend AEP to the multi-robot setting, we combine it with %Dynamic Voronoi Cell (DVC)
DVC~\cite{cortes2004coverage}, which divides the environment into robot-specific spatial regions for decentralized viewpoint selection.

These baselines cover representative exploration strategies, including frontier-based coordination (Hungarian), hierarchical area decomposition (RACER), and information-gain based exploration (AEP). Except for the model ablation section, the default foundation model for the proposed framework is GPT-4o, which is the best performing model from ablation studies. with temperature at 0.2 and max token limit at 1000. For fairness, all methods are evaluated under the same map representation, sensing configuration, robot initialization protocol, and task setup, with $T_{\text{coord}}$ = 50. All methods use the same downstream motion execution and stopping criteria, so that performance differences primarily reflect the high-level coordination strategy rather than low-level control or perception differences.

Existing foundation model based MRS exploration framework such as CoNavGPT~\cite{yu2023co} and COMRES-VLM~\cite{Wang2025COMRESVLMCM} rely on raw visual observations or image-based map inputs as model prompts. As %our method 
SAGR operates on a structured semantic-topological abstraction rather than visual inputs, %these systems 
they are not directly comparable in terms of input representation and computational assumptions, and thus are not included as baselines.

\subsection{Explore and Search}
We first evaluate all methods on both the \textit{exploration} and \textit{semantic target search} tasks across all generated scenarios. Although semantic target search is the primary task considered in this work, pure exploration is also important to evaluate, as SAGR behaves as an exploration strategy when no room instances of the target type have yet been discovered, or when no target room prior is provided. To avoid saturating small environments with excessive robots, which can reduce performance differences between methods, we categorize the apartment layouts into three scales: \textit{small}, \textit{medium}, and \textit{large}. Accordingly, we deploy 2 robots in small layouts, 3 robots in medium layouts, and 4 robots in large layouts.

\begin{table*}[t]
\centering
\caption{Average completion time (in time steps) for exploration and search tasks under different environment scales.}
\label{tab:strategy_comparison}
\resizebox{\textwidth}{!}{
\begin{tabular}{lcccccc}
\toprule
\multirow{2}{*}{\textbf{Strategy}} 
& \multicolumn{2}{c}{\textbf{Small}} 
& \multicolumn{2}{c}{\textbf{Medium}} 
& \multicolumn{2}{c}{\textbf{Large}} \\
\cmidrule(lr){2-3} \cmidrule(lr){4-5} \cmidrule(lr){6-7}
& \textbf{Explore} & \textbf{Search} 
& \textbf{Explore} & \textbf{Search} 
& \textbf{Explore} & \textbf{Search} \\
\midrule
RACER      & $585.1 \pm 154.6 $ & $265.4\pm 194.7$ & $733.0 \pm 166.0$ & $283.9\pm 194.9 $ & $1129.3 \pm 263.6$ & $508.8 \pm 436.2$ \\
Hungarian  & $\mathbf{474.2\pm 70.8}$ & $202.9 \pm 148.8$ & $536.6 \pm 101.3$ & $245.8\pm 206.6$ & $800.1 \pm 117.0$ & $282.1 \pm 210.4$ \\
AEP + DVC  & $488.8 \pm 89.8$ & $196.6 \pm 142.3$ & $ \mathbf{517.2 \pm 84.3}$ & $ 224.4 \pm 194.0$ & $\mathbf{738.6 \pm 102.9}$ & $280.8 \pm 253.7$ \\
SAGR (ours)     & $500.5\pm 95.3$ & $\mathbf{178.8 \pm 141.0}$ 
           & $559.9\pm 78.2$ & $\mathbf{190.4\pm 163.7}$ 
           & $825.6\pm 108.6$ & $\mathbf{228.0 \pm 177.4}$ \\
\bottomrule
\end{tabular}
}
\end{table*}

Table~\ref{tab:strategy_comparison} reports the average task completion time (in time steps) for both exploration and search tasks across different environment scales. For pure exploration, specialized geometric coordination strategies remain the strongest baselines, as expected since they are designed to optimize coverage efficiency directly from frontier geometry. In small environments with two robots, the classical Hungarian frontier assignment performs best due to its simplicity and efficient distance-based allocation. As the environment size and team size increase, AEP + DVC begins to outperform Hungarian, benefiting from its %Dynamic Voronoi Cell (DVC) 
DVC coordination mechanism, which partitions the workspace among robots and reduces redundant exploration. Nevertheless, SAGR remains competitive across all environment scales despite reasoning at a higher semantic level.

In particular, RACER exhibits consistently weaker performance across exploration scenarios. While RACER employs HGrid decomposition with CVRP-based task assignment, coordination is limited to pairwise interactions and does not maintain persistent memory of grid allocations. Hence, grid assignments may change across replanning cycles, leading to inconsistent coverage and reduced coordination efficiency, particularly in larger environments with more robots.

SAGR is not designed to outperform purely geometric methods on exploration-only objectives; rather, it is designed to \emph{preserve strong exploration efficiency while enabling task-aware coordination for semantic search} when natural language guidance, such as, find my jacket left in my bedroom, is available. This is reflected in Table~\ref{tab:strategy_comparison}. At the beginning of a semantic search task, the target room type is typically not yet present in the semantic area graph, so SAGR initially coordinates robots in a manner similar to standard exploration strategies. As new rooms are discovered and semantic labels become available, the LLM dynamically shifts coordination toward rooms whose semantic type matches the target description, while still maintaining exploration to account for the possibility of multiple candidate rooms. This allows robots to prioritize likely target regions without requiring prior knowledge of room locations, leading to consistently faster semantic search %than 
compared to purely geometric coordination strategies. 

SAGR naturally transitions from general exploration to targeted semantic search as task-relevant structure becomes available. The semantic-search gains become more pronounced as the environment scale increases. In large environments, SAGR reduces average search completion time by approximately \textbf{19.2\%} relative to Hungarian assignment and \textbf{18.8\%} relative to AEP + DVC, while remaining in the same overall performance range for exploration.

A common concern %when using 
with the use of LLMs for %multi-robot 
MRS coordination is whether assignments can be generated in near real time and whether the resulting plans remain stable across runs due to the stochastic nature of LLM inference. To evaluate these aspects, we conducted experiments in a medium-scale environment with fixed scenario configurations over 10 runs. The results are reported in Table~\ref{tab:overall_comparison}. Deterministic baselines such as Hungarian and RACER exhibit zero variance in completion steps. In contrast, SAGR introduces stochasticity due to the LLM-based planning process. However, the observed variance remains comparable to AEP + DVC, indicating that LLM-based coordination remains stable across runs. 

In terms of computational cost, the semantic area graph provides a highly structured representation of the environment, allowing compact prompts with minimal complementary information. As a result, SAGR requires approximately $2.5~s$ per coordination query, %which is 
comparable to the RACER baseline using HGrid decomposition. In practice, each query contains fewer than 500 tokens, enabling efficient LLM inference while maintaining near real-time coordination performance. The benefit of the structured semantic area graph is further demonstrated in the ablation study (next subsection), where the foundation model can run locally with near real-time.

\begin{table}[t]
\centering
\setlength{\tabcolsep}{3.5pt}  % default is 6pt
\caption{Task performance and computational cost comparison.}
\label{tab:overall_comparison}
\begin{tabular}{lccc}
\toprule
\textbf{Strategy} 
& \textbf{Explore} 
& \textbf{Search} 
& \textbf{Compute (s)} \\
\midrule
RACER     
& $796.0 \pm 0.0$ 
& $315.0 \pm 0.0$ 
& $2.092 \pm 0.005$ \\

Hungarian  
& $503.0 \pm 0.0$ 
& $363.0 \pm 0.0$ 
& $\mathbf{0.032 \pm 0.004}$ \\

AEP + DVC  
& $\mathbf{479.8 \pm 48.5}$ 
& $286.1 \pm 97.1$ 
& $0.613 \pm 0.100$ \\

SAGR (ours)      
& $537.8 \pm 54.5$ 
& $\mathbf{207.1 \pm 53.1}$ 
& $2.531 \pm 0.392$ \\

\bottomrule
\end{tabular}
\end{table}

\subsection{Ablation Study}
We conduct two ablation studies to analyze the contributions of the key SAGR components. % of the proposed framework.

The first ablation study evaluates how different contextual information provided to the LLM affects coordination performance. In the full SAGR framework, the LLM receives structured information including room connectivity, frontier statistics, task specification, and a brief plan summary describing the intended coordination strategy. To evaluate their importance, we remove each component individually and measure the resulting search performance. 

\begin{table}[t]
\centering
\caption{Ablation study on SAGR components.}
\label{tab:ablation_info}
\begin{tabular}{lcc}
\toprule
\textbf{Variant} & \textbf{Time (steps)} & \textbf{Change} \\
\midrule
Full SAGR 
& $\mathbf{189.0 \pm 142.9}$ & -- \\
\midrule
-- No Neighborhood Context 
& $207.3 \pm 168.1$ & $\uparrow +9.6\%$ \\
-- No Plan Summary 
& $213.7 \pm 172.9$ & $\uparrow +13.0\%$ \\
-- No Target Room Type 
& $219.2 \pm 162.7$ & $\uparrow +16.0\%$ \\
\bottomrule
\end{tabular}
\end{table}

Table~\ref{tab:ablation_info} reports the average completion time for the semantic search task. Removing any component consistently degrades performance. In particular, removing information about target room types, e.g., \emph{search for the jacket I left in the bedroom}, results in the largest performance drop (+16.0\%), indicating that explicitly specifying the search objective is critical for effective room prioritization. Removing the plan summary also degrades performance (+13.0\%), as it prevents the LLM from maintaining consistent coordination across planning cycles. Eliminating neighborhood connectivity information further reduces performance (+9.6\%), suggesting that spatial context helps the LLM distribute robots more effectively across rooms. Overall, these results highlight the importance of structured contextual information for reliable LLM-based coordination.

The second ablation study evaluates how the choice of foundational LLM affects the performance of the proposed framework. We compare both cloud-based models accessed through API calls and locally deployed open-source models running on a server with four NVIDIA A5000 GPUs.

\begin{table}[t]
\centering
\caption{Effect of different LLMs on search performance.}
\label{tab:llm_models}
\begin{tabular}{lcc}
\toprule
\textbf{Model} & \textbf{Time (steps)} & \textbf{Compute (s)} \\
\midrule
\multicolumn{3}{c}{\textit{Cloud Models}} \\
\midrule
GPT-4o 
& $\mathbf{192.3\pm 161.7}$ 
& $2.091 \pm 0.400$ \\
Claude-Haiku-4.5 
& $197.8 \pm 183.8$ 
& $2.911 \pm 0.500$ \\
Gemini-2.5-Flash-Lite 
& $232.2 \pm 193.5$ 
& $1.403 \pm 0.218$ \\
\midrule
\multicolumn{3}{c}{\textit{Local Models}} \\
\midrule
Qwen2.5-7B-Instruct 
& $223.6 \pm 188.3$ 
& $6.298 \pm 1.643$ \\
Meta-Llama-3.1-8B-Instruct 
& $224.9 \pm 185.5$ 
& $5.747 \pm 0.522$ \\
DeepSeek-R1-Distill-Llama-8B 
& $255.6 \pm 213.8$ 
& $5.725 \pm 1.909$ \\
\bottomrule
\end{tabular}
\end{table}

The results are summarized in Table~\ref{tab:llm_models}. Among cloud-based models, GPT-4o achieves the best performance, while Gemini-2.5-Flash-Lite provides faster inference at the cost of slightly degraded coordination quality. For locally deployed models, Qwen2.5-7B and Llama-3.1-8B achieve competitive performance but exhibit higher inference latency compared to cloud models. Overall, these results demonstrate that SAGR is compatible with a wide range of LLM backbones and can be deployed with both cloud-based and locally hosted models depending on computational constraints.

\section{Conclusion}
This paper introduced \textit{Semantic Area Graph Reasoning (SAGR)}, a hierarchical framework for multi-robot exploration and semantic target search in %previously 
unknown indoor environments. By constructing a semantic area graph from an incrementally built semantic occupancy map, SAGR enables coordination at the room level rather than the traditional frontier-cell level. This compact abstraction %allows 
enables %a large language model 
an LLM to reason over the discovered semantic structure of the environment and assign robots to candidate regions according to the task objective.

The proposed framework integrates LLM-based high-level reasoning with deterministic frontier selection and local motion execution. This design preserves the efficiency and reliability of classical exploration algorithms while enabling semantic task information to directly influence multi-robot coordination. Experimental results on realistic apartment environments show that, although specialized geometric methods remain strong baselines for pure exploration, SAGR maintains competitive exploration performance while substantially improving efficiency on semantic search tasks. %The framework also maintains practical computational performance, with 
Moreover, the compact coordination prompts and inference times are suitable for near real-time deployment.

Future work will investigate scaling semantic reasoning to larger robot teams and more complex environments. In particular, hierarchical coordination strategies involving multiple LLM agents may enable more scalable task decomposition and distributed decision-making. Another promising direction is to incorporate longer horizon planning directly over the semantic area graph. Recent advances such as Graph-of-Thought~\cite{besta2024graph} reasoning align naturally with our semantic area graph representation and may allow the LLM to perform structured multi-step reasoning. This could enable the system to plan multi stage search strategies and better exploit the semantic area graph for task-aware coordination.

\section*{Acknowledgments}

This work is sponsored in part by the AFOSR under the award number FA9550-19-1-0169, and by the NSF under NAIAD Award 2332744 as well as the National AI Institute for Edge Computing Leveraging Next Generation Wireless Networks, Grant CNS-2112562.

\bibliographystyle{IEEEtran}
\bibliography{bib}
\end{document}